\documentclass[lettersize,journal]{IEEEtran}
\usepackage{amsmath,amsfonts}
\usepackage{algorithmic}
\usepackage{algorithm}
\usepackage{array}
\usepackage[caption=false,font=normalsize,labelfont=sf,textfont=sf]{subfig}
\usepackage{textcomp}
\usepackage{stfloats}
\usepackage{url}
\usepackage{verbatim}
\usepackage{graphicx}
\usepackage{cite}
\usepackage{times}
\usepackage{epsfig}
\usepackage{graphicx}
\usepackage{amsmath}
\usepackage{amssymb}

\usepackage{tabu}
\usepackage{booktabs}
\usepackage{mathtools}
\usepackage{wrapfig}
\usepackage{textcomp}
\usepackage{bbm}
\usepackage{multirow}
\usepackage{float}

\DeclareRobustCommand\onedot{\futurelet\@let@token\@onedot}
\def\@onedot{\ifx\@let@token.\else.\null\fi\xspace}

\def\ie{\emph{i.e}\onedot}

\def\etal{\emph{et al}\onedot}

\usepackage{caption} 
\begin{document}

\title{PONet: Robust 3D Human Pose Estimation via Learning Orientations Only}

\author{Jue Wang, Shaoli Huang{\textsuperscript{*}},Xinchao Wang, Dacheng Tao
   \thanks{{\textsuperscript{*}}Corresponding authors.}
   }

\markboth{Journal of \LaTeX\ Class Files,~Vol.~14, No.~8, August~2021}%
{Shell \MakeLowercase{\textit{et al.}}: A Sample Article Using IEEEtran.cls for IEEE Journals}

\IEEEpubid{0000--0000/00\$00.00~\copyright~2021 IEEE}

\maketitle

\begin{abstract}
    Conventional 3D human pose estimation
    relies on first detecting 2D body keypoints 
    and then solving the 2D to 3D correspondence problem.
    Despite the promising results,
    this learning paradigm is highly dependent on the 
    quality of the 2D keypoint detector, 
    which is inevitably 
    fragile to occlusions and out-of-image absences.
    In this paper,
    we propose a novel {Pose Orientation Net~(PONet)}
    that is able to
    robustly estimate 3D pose by learning orientations only,
    hence bypassing the error-prone keypoint detector
    in the absence of image evidence.
    For images with partially invisible limbs, PONet  estimates the 3D orientation of these limbs by taking advantage of the local image evidence to recover the 3D pose.
    Moreover, PONet is competent to infer full 3D poses even from images with completely invisible limbs, by exploiting the orientation correlation between visible limbs to complement the estimated poses,
    further improving the robustness of 3D pose estimation.
    We evaluate our method on multiple datasets, including Human3.6M, MPII, MPI-INF-3DHP, and 3DPW.
    Our method achieves results on par with state-of-the-art techniques in ideal settings, yet significantly eliminates the dependency on keypoint detectors and the corresponding computation burden.
    In highly challenging scenarios, such as truncation and erasing,
    our method performs very robustly and yields
    {much superior results}
    as compared to  state of the art,
    demonstrating its potential for real-world applications.
    Our code will be made publicly available.
    Please see our video results
    in the supplementary materials
    for pose {estimation in extreme cases}.
\end{abstract}


\section{Introduction}

\begin{figure}[htb]
    \centering
    \includegraphics[width=1\linewidth]{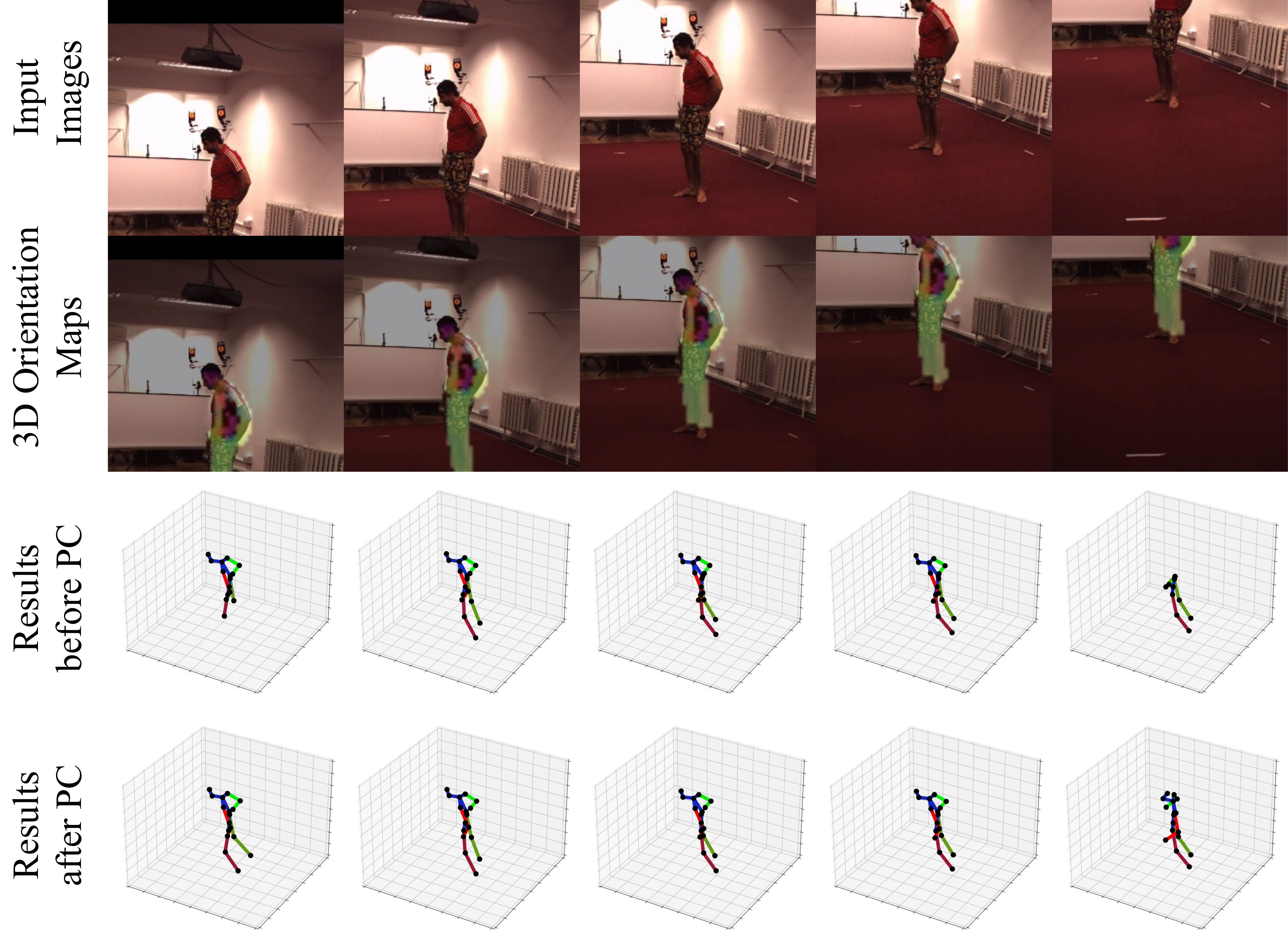}
    \caption{Example results on Human3.6M with vertical translation from $-40\%$ to $40\%$. PC denotes pose complementation. Our PONet is very robust to images with invisible limbs. More visual results on challenging scenarios can be found in our supplement.}
    \label{fig:title}
\end{figure}

\IEEEPARstart{H}uman pose estimation aims at recovering the coordinates of a human body  from one or multiple images, and therefore plays a vital role in an exceptionally broad spectrum of applications. 
State-of-the-art 3D human pose estimation methods~\cite{zhou2016sparseness,moreno20173d,Chen_2017_CVPR,martinez2017simple,lee_2018_eccv,Kanazawa_2019_CVPR,Li_2020_CVPR} rely on first detecting several 2D keypoints, like the body joints, from the image, followed by mapping the 2D keypoint locations back to the 3D world.   
The advantage of building 3D pose estimation on 2D keypoint detection 
lies in
that the former can inherit the good generalization capacity from the latter.
Despite its popularity,
this scheme still suffers from several flaws.
Firstly, methods built on 2D keypoint detection are usually sensitive to the 2D detection errors,
since the 2D-to-3D mapping is a highly 
ill-posed and -conditioned problem.
Minor errors in the locations of the 2D keypoints 
may lead to major drifts in the 3D results.
Secondly, 2D keypoint detection requires all the body joints to be visible in the image. 
Such prerequisite is justifiable for monitored lab environments, 
but unfortunately too strong for practical application scenarios, 
where out-of-image absences of body joints and heavy occlusions 
frequently occur and thus deteriorate the 3D estimation results.

Some recent endeavors have thus focused on
bypassing 2D keypoint detection through recovering human mesh.
For instance, Kanazawa~\etal~\cite{Kanazawa_2018_CVPR} propose a method that directly estimates the parameters of  SMPL~\cite{loper2015smpl}
and then infer the 3D pose from the 3D body shape. 
These methods provide richer 3D information about the body, like 3D human skin, while satisfying the anthropometric constrains.
Although this method is capable to handle keypoint-absent cases, 
it would fail if a whole limb~\footnote{We denote a limb to be the  body region between two adjacent joints.} is visually absent.


In this paper, we attempt to study
3D human pose estimation in challenging
scenarios, where image evidences are incomplete.
To this end, we first 
categorize such evidence-incomplete images
into two levels of difficulties:
images with only \emph{missing joints},
\ie each limb is at least partially visible,
and images with \emph{missing limbs},
\ie at least one limb is completely invisible.
The latter case is unarguably more demanding
than the former, since more visual cues
are absent. However, state-of-the-art methods
and even benchmarks have been focused on complete images or
the former case only, yet have largely overlooked
the latter, which, unfortunately, 
better reflects in-the-wild data 
in real-world applications.

We propose here a robust 3D pose estimation
termed as \emph{Pose Orientation Net}~(PONet),
which allows us to effortlessly 
to handle both scenarios,
as demonstrated in  Fig.~\ref{fig:title}.
{At the heart of our approach is region-based 3D orientation learning and pose complementation.}
Specifically, our method estimates limb confidence maps, 2D and 3D limb orientation maps directly from images using a three-branch multi-stage fully convolutional neural network~(FCNN).
The limb confidence map represents the probability of each pixel  
within the corresponding limb region,
while the 2D and 3D orientation maps represent
the limb orientations in the pixel space and 3D space, respectively.
The 3D pose estimation is produced by integrating the estimated 3D limb orientations with an embedded skeleton-shaped human model, which consists of free joints and fixed-length limbs.
This region-based orientation learning makes 
our method robust to the cases where  keypoints are missing.
{
The motivation behind such a design lies in that, 
3D limb orientations enable 3D human pose recovery by utilizing the estimated directional vector oriented from a neighboring body part, 
even if the body part of interest is visually incomplete.}

The proposed method also includes a pose complementation sub-network,
which aims to estimate a full 3D pose 
when some limbs are missing.
The idea is to infer the 3D configuration of missing body parts from the visible parts, using the prediction confidence as an indicator.
Intuitively, the correlation between 3D limb orientations is stronger than that between keypoints, making it more dependable to infer a full pose 
based on orientation relations between visible limbs.
We thereby infer the 3D orientations of invisible limbs from initial 3D orientation predictions of visible limbs,
by taking 3D orientation correlation and limb confidence into account.
The pose complementation sub-network further improves the robustness of 3D pose estimation, especially on images with inaccurate bounding boxes and large translation.

Unlike prior orientation-based methods that usually combine orientation 
learning with 2D keypoint detection, 
all the three maps adopted in our method are limb-region-based representation.
Such a design not only allows us to
handily carry out knowledge transfer
to images without 3D annotations
by using limb region estimation and 2D orientation learning, which can be trained with both 2D and 3D data, as a bridge,
but also enable differentiable post-processing and end-to-end training.
Besides, we take the average 3D orientation vector, weighted by the region confidence of each pixel,
as the final estimated 3D limb orientation. This pixel-wise voting scheme reduces the effect of noise and outliers in the estimated 3D orientation maps, making the 3D orientation estimation and pose estimation more robust and stable.

We evaluate our method on several benchmarks, including Human3.6M, MPII, MPI-INF-3DHP, and 3DPW. Our proposed method achieves performance on par with state-of-the-art techniques in standard settings,
but shows {much stronger} generalization capacity to unseen in-the-wild images.  
Moreover, unlike prior methods that usually rely on external keypoint detectors~\cite{martinez2017simple,moreno20173d,lee_2018_eccv,Cheng_2019_ICCV,Qiu_2019_ICCV,Li_2020_CVPR} or complex human models~\cite{bogo2016keep,Kanazawa_2018_CVPR,Pavlakos_2019_CVPR,Zhang_2020_CVPR}, 
our method stands on its own
and 
does not require any other third-party pre-trained model. 
Most importantly, our method significantly outperforms state-of-the-art techniques on
robustness testing including translation, occlusion, and various erasing, 
demonstrating its potential for real-world applications.

Our main contributions are summarized as follows.
%
%
\begin{itemize}
    \item We propose PONet for robust 3D human pose estimation, which is capable to infer the complete 3D pose even when the input image is incomplete. Our method significantly outperforms state-of-the-art techniques in challenging scenarios such as heavy object-occlusion, large translation and various erasing. 
    \item We propose a region-based orientation learning, which allows us to handily carry out knowledge transfer to in-the-wild images without 3D annotations. Our method yields very promising results on MPI-INF-3DHP and 3DPW without training on them, validating its gratifying generalization capacity.
    \item We propose a differentiable voting scheme to extract orientations from predicted orientation maps, allowing us to end-to-end train the network and making the orientation estimation more robust and stable.
\end{itemize}

\section{Related work}
In this section, we briefly review monocular 3D human pose estimation approaches by dividing them into three categories that may overlap each other, \ie 2D keypoint detection based, human mesh recovery based and orientation-based approaches.

\textbf{2D keypoint detection based.}
2D human pose estimation has witnessed unprecedented progress in recent years and lots of 3D pose estimation methods are built on this technique.
Early efforts on 3D pose estimation used dictionary learning, with the assumption that a 3D pose can be represented by a sparse linear combination of a set of basis poses~
\cite{Wang_2014_CVPR,Zhou_2015_CVPR,zhou2016sparseness,zhou2017sparse,zhou2018monocap}. 
Recently, more and more researchers start to use deep neural networks for 3D pose regression~\cite{martinez2017simple,Sun_2017_ICCV,Tome_2017_CVPR,moreno20173d,nie2017monocular,fang2018learning,lee_2018_eccv,Kocabas_2019_CVPR,Arnab_2019_CVPR,Li_2020_CVPR}. 
For instance, 
Martinez~\etal~\cite{martinez2017simple} propose a light weighted fully connected network with residual connections and achieved impressing results.
Lee~\etal~\cite{lee_2018_eccv} propose a long short-term memory (LSTM) architecture to reconstruct 3D depth from the centroid to edge joints through learning the joint inter-dependencies.
Li~\etal~\cite{Li_2020_CVPR} improves 2D-to-3D regression by synthesizing new 2D-3D pairs with evolution algorithm.
Cheng~\etal~\cite{Cheng20203DHP} use explicit occlusion augmentation to improve the robustness to keypoint detection and 3D pose estimation in image sequences. 
The major problem of keypoint-detection-based approaches is that 2D-to-3D lifting is ill-posed and ill-conditioned.
Minor errors in the locations of 2D keypoints can have large consequences in the 3D results.
In addition, these methods are prone to ambiguities in the 2D to 3D correspondence step. 

\textbf{Human mesh recovery based.}
Human mesh recovery is a highly related topic with 3D human pose estimation.
Several recent works deal with both tasks.
These methods~\cite{bogo2016keep,lassner2017unite,tung2017self,Kanazawa_2018_CVPR,pavlakos2018learning,Arnab_2019_CVPR,Varol_2018_ECCV,Kanazawa_2019_CVPR} recover the 3D human body shape
based on the generative  body model SMPL~\cite{loper2015smpl}.
Bogo~\etal~\cite{bogo2016keep} propose SMPLify, an optimization-based method to recover SMPL parameters from detected 2D joints that leverages multiple priors.
Lassner~\etal~\cite{lassner2017unite} take curated results from SMPLify to train 91 keypoint detectors, some of which correspond to the traditional body joints and others correspond to locations on the surface of the body.
Kanazawa~\etal~\cite{Kanazawa_2018_CVPR} directly infer SMPL parameters from images without relying on detected 2D keypoints, allowing shape and pose estimation on images with missing joints.
However, these methods would fail on images with missing limbs.

\textbf{Orientation learning based}
There are a handful of approaches try to learn orientations for 3D pose estimation.
Zhou~\etal~\cite{zhou2016deep} propose a kinematic human model that adds different constrains to different joints, to simulate real human structure.
They use CNN to learn rotation angles for the adjustable joints.
Cao~\etal~\cite{Cao_2017_CVPR} propose \emph{part affinity field}~(PAF) to help linking the keypoints on a person in the multi-person 2D pose estimation problem.
Xiang~\etal~\cite{Xiang_2019_CVPR} use PAFs to fit a deformable mesh model to recover 3D shape and pose. 
Luo~\etal~\cite{luo2018orinet} introduce OriNet that predicts 2D keypoint heatmaps and 3D PAFs for 3D pose estimation.
In Liu~\etal's work~\cite{liu2019improving}, 3D orientations are used as additional image evidence to improve the 2D-to-3D regression.
Shi~\etal~\cite{10.1145/3407659} predict 3D pose by estimates 3D orientations and limb lengths from 2D keypoints. 
Again, these approaches rely heavily on 2D keypoints. As a result, they are fragile to keypoint absences as well, let alone cases with limb absence. 

\textbf{Our approach.} 
The proposed approach estimates 3D pose by only learning limb orientations from images without detecting 2D keypoints.
Our region-based orientation learning allows 3D pose estimation on images where the limbs are partially visible.
In addition, the proposed PONet can infer a full body pose from images with completely invisible limbs, by exploiting the orientation correlation between visible limbs.
Compared to traditional keypoint-detection-based and SMPL-based 3D pose estimation approaches, our method eliminates the dependency on 2D keypoint detectors, and is much more robust to images with visually absent joints or limbs.

\begin{figure*}[tbh]
    \centering
    \includegraphics[width=\textwidth]{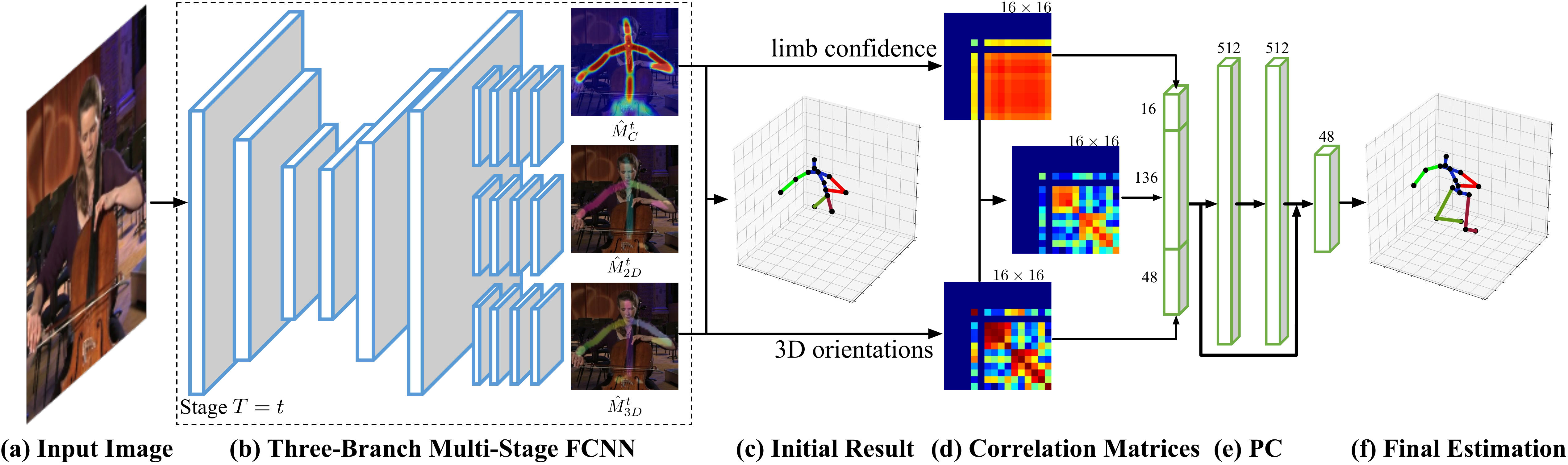}
    \caption{The inference pipeline of PONet. The system takes as input a color image, which can be incomplete.
    First, the three-branch multi-stage FCNN simultaneously estimates limb confidence maps $\hat{M}_c$ and 2D/3D limb orientation maps $\hat{M}_{2D}$ and $\hat{M}_{3D}$.
    Then we extract 3D limb orientations from the predicted maps using a differentiable voting scheme to produce an initial 3D pose estimation.
    Finally, we use a pose complementation~(PC) sub-network, to infer a complete and reasonable final 3D pose estimation from initial estimation.
    }
    \label{fig:pipeline}
\end{figure*}

\section{Method}
    The inference pipeline of the propose PONet is illustrated in Fig.~\ref{fig:pipeline}.
    Our method takes as input a single color image~(Fig.~\ref{fig:pipeline}a), which can be incomplete and have visually absent joints or limbs.
    Firstly, we use a three-branch multi-stage fully convolutional neural network~(Fig.~\ref{fig:pipeline}b) to simultaneously predict three sets of maps, \ie the limb confidence maps $\hat{M_C}$, 2D limb orientation maps $\hat{M}_{2D}$ and 3D orientation maps $\hat{M}_{3D}$.
    Then we use a differentiable voting scheme to extract 3D limb orientations from the predicted maps.
    An initial 3D pose estimation~(Fig.~\ref{fig:pipeline}c) is produced by integrating the predicted 3D limb orientations into  a skeleton-shaped human model with fixed limb lengths.
    Since the input image may be incomplete, we introduce a pose complementation sub-network~(Fig.~\ref{fig:pipeline}e) to infer the 3D orientations of the missing limbs from that of the visible ones, so that the network can estimate a complete final 3D pose~(Fig.\ref{fig:pipeline}f) even if the input image is incomplete.
    In the following sections, we explain these steps in detail.

    
\subsection{Collaborative orientation learning}
    \subsubsection{Limb orientation representation} \label{sec:ori_rep}
        Following~\cite{Cao_2017_CVPR}, we use orientation maps to represent the limb orientations.
        Specifically, given an image of size $w\times h$, a 3D orientation map is a $w'\times h'$  three-channel map, where each pixel within the limb region is a directional vector representing the 3D orientation of the limb in 3D space.
        The pixels outside the limb region are set to zero vectors.
        The 2D orientation maps are defined similarly as 3D orientation maps.
        The limb confidence map is a 1-channel map, where each pixel is a value between 0 and 1, representing the probability of the pixel within the limb region.
    
        In human pose estimation, the exact human limb region is unavailable.
        For simplicity, we take the rectangular region defined by the two end points of the limb and a fixed width $d$ as the limb region.
        When the 2D length of the limb is smaller than $d$, we take the square region centred at the mid-point of the limb as the limb region,
        to avoid a too small region while keeping the 2D orientation information.
        In this work, the 3D pose consists of $17$ joints and thus $16$ limbs.
        The FCNN predicts these three sets of maps simultaneously, that is $16\times(1+2+3)=96$ channels in total. 
    
    \subsubsection{Weakly-supervised 3D orientation learning}\label{sec:ori_learning}
        Prior methods usually require specially designed losses~\cite{zhou2017towards,Dabral_2018_ECCV}, additional annotations~\cite{Pavlakos_2018_CVPR} or adversarial training~\cite{Yang_2018_CVPR} to achieve knowledge transfer to in-the-wild images.
        By contrast, our PONet can handily generalize to in-the-wild images without extra bells and whistles but achieving better generalization capacity.
        Specifically, we use a three-branch multi-stage FCNN to predict the limb confidence maps, 2D and 3D orientation maps, respectively. 
        Estimation of 2D orientation map is an auxiliary task, which is only used in the training phase to help learning orientation collaboratively and improve generalization.
        During training, each mini-batch is randomly sampled from two datasets, an in-the-wild 2D pose dataset MPII~\cite{andriluka20142d} and an indoor 3D pose dataset Human3.6M~\cite{ionescu2014human3}, with equal probability.
        Since both datasets have 2D annotations, we can fully-supervised train the limb confidence map prediction branch and 2D orientation map prediction branch.
        For the 3D orientation branch, on average only a half of the training examples, \ie those sampled from the 3D dataset, have 3D orientation map supervisions.
        The other half of the training data are left unsupervised in this branch.
        Our experiments demonstrate that, the proposed PONet can easily generalize to in-the-wild images using this simple weakly-supervised training strategy and achieve much better generalization capacity than state-of-the-art methods.
      
        In this paper, the confidence map estimation is taken to be a pixel-wise binary classification problem.
        We use the \emph{Binary Cross Entropy}~(BCE) loss for this task:
        \begin{equation}
        \mathcal{L}_{\text{CM}}^t=\text{BCE}(\hat{M}^t_{\text{C}},{M}^t_{\text{C}}),
        \end{equation}
         where $\hat{M}^t_{\text{C}}$ and ${M}^t_{\text{C}}$ represent the predicted and ground truth limb confidence maps at stage $t$, respectively. 
         The 2D and 3D orientation map estimation is taken to be a regression problem and we use \emph{Mean Squared Error}~(MSE) as the loss function:
        \begin{equation}
            \mathcal{L}_{\text{OM2D}}^t=\text{MSE}(\hat{M}^t_{\text{2D}}, {M}^t_{\text{2D}}),
        \end{equation}
         \begin{equation}
            \mathcal{L}_{\text{OM3D}}^t=\text{MSE}(\hat{M}^t_{\text{3D}}, {M}^t_{\text{3D}}),
        \end{equation}
         where $\hat{M}^t_{\text{2D/3D}}$ and ${M}^t_{\text{2D/3D}}$ denote the predicted and ground truth 2D/3D orientation maps at stage $t$, respectively. 
    
    \subsubsection{Differentiable voting scheme for orientation extraction} \label{sec:ori_extract}
    Prior orientation based methods~\cite{Cao_2017_CVPR,luo2018orinet} extract the directional vectors from orientation maps by computing the line integral over the corresponding orientation maps along the line segment connecting the detected part locations.
    Many other candidate directional vectors that not on the line segment, which also contain important orientation information, however, are discarded.
    Besides, this procedure depends on the detected 2D keypoint locations, making it non-differentiable and thus disables end-to-end training.
    
    To address the above problems, we present a differentiable voting scheme to extract orientation vectors from the predicted orientation maps.
    As described in Section~\ref{sec:ori_rep}, our method predicts the limb confidence maps and 3D orientation maps at the same time.
    For each limb $i$ , we extract its 3D limb orientation by taking average of all the pixels in the pixel-wise product of the predicted limb confidence map $\hat{M}_{\text{C}i}$ and 3D orientation map $\hat{M}_{\text{3D}i}$ of the last stage:
    \begin{equation}\label{eq:vote}
        v_i = \frac{1}{w'h'}\sum_{w'}\sum_{h'}\hat{M}_{\text{C}i}\odot \hat{M}_{\text{3D}i},
    \end{equation}
    where $\odot$ denotes pixel-wise product. Obviously, Eq.~\ref{eq:vote} is differentiable, which makes end-to-end training possible.
    Besides, this confidence weighted voting scheme can leverage all the predicted orientation information, making our orientation estimation more accurate and stable.

    \subsubsection{Integration 3D orientations with human model}\label{sec:integration}
    Our method generates 3D pose estimation by integrating the estimated 3D orientations with a skeleton-shaped human model with fixed-length limbs.
    This process works like twisting the limbs of the human model to fit the predicted 3D limb orientations.
    Like many prior methods, we treat the pelvis as the root node and fix it to the origin.
    For each child node $i$, its location $Y_i$ is determined by its parent node's location $X_i$, the estimated 3D orientation $v_i$ and the 3D limb length $L_i$ as follows:
    \begin{equation} \label{equ:integration}
        Y_i = X_i + L_i v_i,
    \end{equation}
    where $X_0=0$. The 3D pose is generated from root node to leaf nodes by recursively applying Eq.~\ref{equ:integration} until all the leaf nodes are determined.
    
    \subsubsection{End-to-end training with 3D pose loss} \label{sec:end2end}
    In Section~\ref{sec:ori_learning}, each limb orientation is \emph{separately} optimized.
    Although limb orientation errors do not accumulate,
    3D joint location errors could propagate along the skeleton tree and possibly accumulate into large errors for joints at the leaf node.
    
    To this end, long-term objectives should be considered so that the 3D orientations are jointly optimized. 
    In our method, since all the steps are differentiable,
    we can directly use the 3D pose loss as the long-term objective and train the model end-to-end.
    This works like slightly adjusting the limb orientation so that the joint locations fit the ground truth better.
    In experiment, we find that end-to-end training can speed up the convergence, and improve the inference accuracy as well.
    Here we use L1 loss for 3D pose following many prior works:
     \begin{equation}
            \mathcal{L}_{\text{P3D}}=|\hat{S}-{S}|,
    \end{equation}
    where $\hat{S}$ and ${S}$ represent the predicted and ground truth 3D pose at the last stage, respectively. 

    \subsection{Human pose complementation}
    The orientation learning network can produce a complete pose estimation unless all limbs are at least partially visible in the image.
    If a limb is completely visually absent, the estimated 3D orientation of the missing limb could be a zero vector or noisy values, resulting in an incomplete or wrong estimation.
    To this end, we propose a pose complementation~(PC) network~(see Fig.~\ref{fig:pipeline}e) to infer the 3D orientations of the invisible limbs from visible ones.
    To achieve this, there are two key problems to answer.
    
    \textbf{(a) Which limbs should be complemented?}
    Different testing images may have different missing limbs and different number of missing limbs.
    How to let the network know which limbs need to be complemented?
    Our solution is straightforward, \ie using the predicted limb confidence score as an indicator.
    Generally, the 3D orientation estimations of invisible limbs tend to have low confidence scores while visible ones have high ones.
    In other words, limbs with lower confidence scores are probably invisible and the corresponding 3D orientations  may need to be re-inferred.
    So we explicitly provide the PC sub-network information about which limbs need complementation by feeding the limb confidence scores into it.
    
\begin{figure}[tbh]
            \centering
            \includegraphics[width=\linewidth]{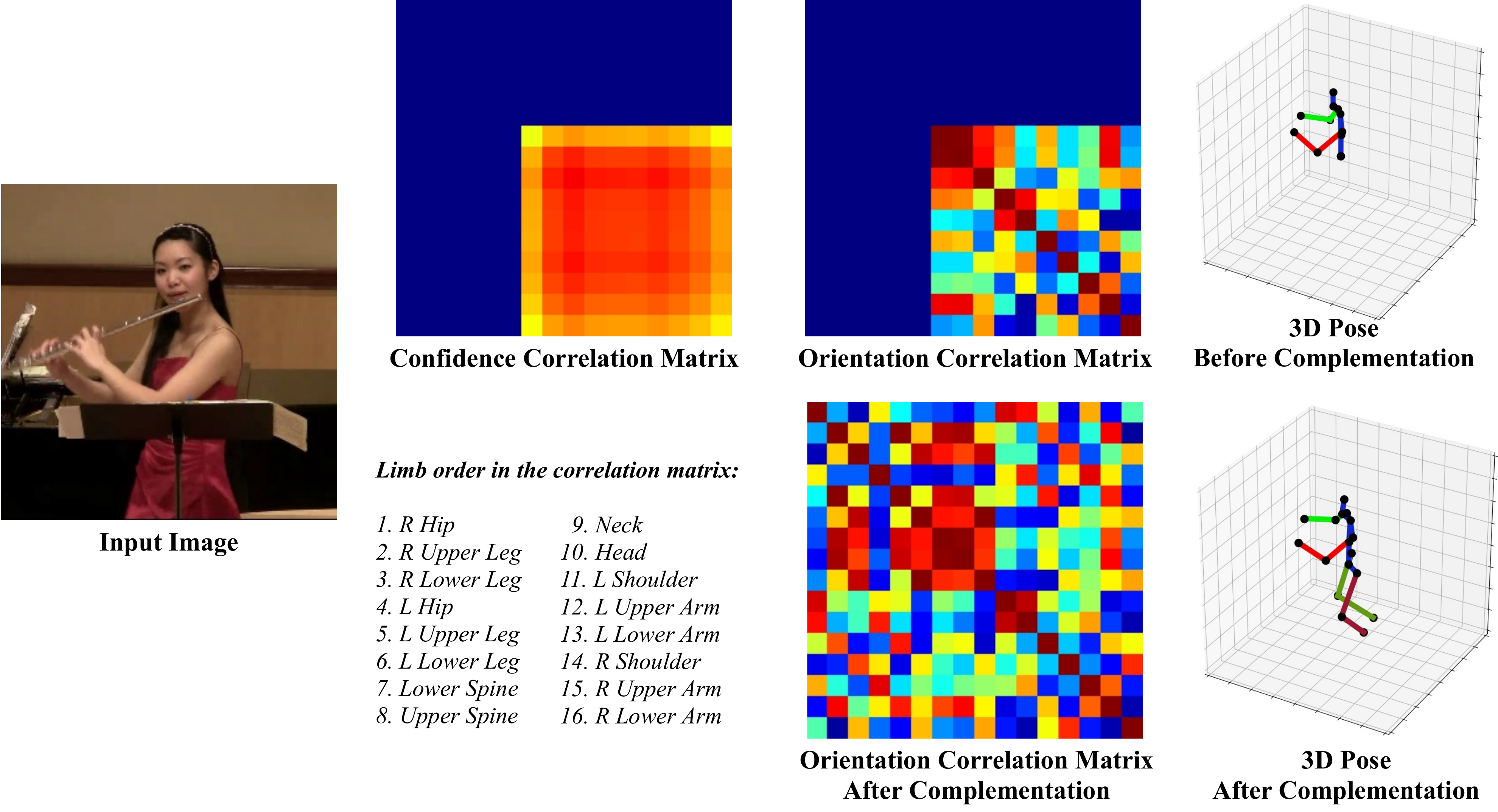}
            \caption{An illustration of the proposed pose complementation.}
            \label{fig:complementation}
        \end{figure}

    \textbf{(b) How to complement?}
    Fig.~\ref{fig:complementation} shows an example of the proposed pose complementation, which is aimed to infer the configuration of missing limbs and provide an overall reasonable pose estimation, while keeping estimation of the visible parts consistent with the input image.
    To achieve this, 
    we feed into the PC sub-network with three components, the predicted limb confidence score, the initial 3D limb orientation estimation and the upper triangular part of the 3D orientation correlation matrix masked by the confidence correlation matrix. 
    The dimension of the input is $16+16\times3 + 16(16+1)/2=200$.
        
    The architecture of the PC sub-network is quite simple, which only consists of two linear layers of size 512 with residual connection.
    In our experiment, we find that the network learns better pose complementation by predicting modification vectors $\Delta v_i$ than directly predicting $v_i^c$:
        \begin{equation}
            v_i^c = \frac{v_i+\Delta v_i}{||v_i+\Delta v_i||_2}.
        \end{equation}
        
        The complemented 3D orientations are then converted into 3D pose $\hat{S}^c$ and optimized with the L1 loss as well:
        \begin{equation}
                \mathcal{L}_{\text{CP3D}}=|\hat{S}^c-{S}|.
        \end{equation}
    
        In all, for a $T$ stages network, our overall objective is the combination of the five objectives:
        \begin{equation}
            \label{eq:loss}
            \small
            \mathcal{L} = \sum_{t=1}^T(\lambda_1 \mathcal{L}_{\text{CM}}^t + \lambda_2 \mathcal{L}_{\text{OM2D}}^t + \mathbbm{1}\lambda_3\mathcal{L}_{\text{OM3D}}^t) + \mathbbm{1}(\mathcal{L}_{\text{P3D}} + \mathcal{L}_{\text{CP3D}}),
        \end{equation}
        where $\lambda_1$, $\lambda_2$ and $\lambda_3$ control the relative importance of each objective, $\mathbbm{1}$ is an indicator function that is 1 if ground truth 3D is available for an image and 0 otherwise. We set $\lambda_1=0.1$, $\lambda_2=1$ and $\lambda_3=1$ in our experiment.
    
\section{Experiments}
    We provide here details on our experiments, including datasets and implementation details,
    results under regular settings, results on images with missing joints/limbs, cross-dataset evaluations to test the generalization capacity and qualitative results.
    
    \subsection{Datasets}
    \textbf{Human3.6M}~\cite{ionescu2014human3} is a large-scale indoor 3D human pose dataset
    that comprises 3.6 million images and the corresponding 2D pose and 3D pose annotations.
    It features $7$ subjects performing $15$ everyday activities.
    Following the standard protocols, we use S1, S5, S6, S7, S8 for training, and S9, S11 for testing.  We report results in two  metrics, \ie \emph{Mean Per Joint Position Error~(MPJPE)} and \emph{MPJPE after Procrustes Alignment~(PA-MPJPE)}.  The original videos are down-sampled from 50fps to 10fps to reduce redundancy.
    
    \textbf{MPI-INF-3DHP}~\cite{mehta2017monocular} test set consists of 2929 indoor and outdoor images from six subjects performing seven actions. We only use the test set of this dataset to evaluate our method's generalization quantitatively.
    
    \textbf{MPII}~\cite{andriluka20142d} is a widely used benchmark for 2D human pose estimation. It contains 5K in-the-wild images covering a wide range of activities.
    It is annotated with 2D keypoints but no 3D ground truth.
    We use it for weakly supervised training to achieve better generalization.
    
    \textbf{3DPW}~\cite{vonMarcard2018} is a recently proposed 3D poses dataset in the wild.
    It contains 60 video sequences and 3D annotations captured via IMUs.
    For a fair comparison, the evaluation is performed following the same protocol as~\cite{Zhang_2020_CVPR}.
    

    \subsection{Implementation details}
    Our method is implemented with PyTorch~\cite{NEURIPS2019_9015}.
    We train our model in two steps from scratch.
    First, we train the whole network in regular settings.
    Augmentations of random scaling~($1\pm0.25$), random translation~($\pm0.2$),  random rotation~($\pm30^{\circ}$), random horizontal flipping~($p=0.5$) and random color jitter~($1\pm0.2$), are used for both MPII and Human3.6M.
    The network is trained for 100 epochs with a batch size of 12.
    The learning rate is initially set to $2\times10^{-4}$ and decays at 60th epoch by a factor of $0.1$.
    
    In the first training step, the images are almost always complete, because the translation augmentation is limited at a very small range.
    To this end, we need to create incomplete images to train the PC sub-network. 
    In our experiments, we use random translation  up to $\pm50\%$  of the image size and randomly render objects on the 3D training images, to simulate joint and limb absence caused by truncation and occlusion.
    Then we fine-tune network on these synthetic incomplete images for $50$ epochs with the same batch size.
    The learning rate is initially set to $1\times10^{-4}$ and decays at the 30th epoch by a factor of $0.1$.
    We use the RMSProp as the optimizer for all the steps.
    In both training steps, the PC sub-network is detached from the FCNN.
    In other words, the gradient in the PC sub-network does not backpropagate to the FCNN, to avoid the deterioration of orientation learning.
    It takes about 35 hours on two Tesla V100 GPUs with 16 GB memory on each to train a 4-stage model.
    
    \subsection{Rules to generate incomplete images}\label{sec:rules}
In Section~\ref{sec:qr} of the main manuscript, we provided quantitative results on various incomplete images.
Here we explain the rules to generate these incomplete images.

\textbf{Translation}. For an image of the size $a\times a$, we randomly translate the image center by $(x, y)\times a$, where $x,y$ are random numbers drawn from uniform distribution between $[-\tau, \tau]$. We set $\tau=0.25, 0.4$ respectively in our experiments.

\textbf{Synthetic occlusions}. For each image, we add a random number of VOC objects at random locations with random sizes. We directly use the code provided by \cite{sarandi2018robust} to generate occlusions.

\textbf{Rectangular erasing}. Given an image of size $a\times a$, we first select two arbitrary points within the image as the two mid-points of the widths of the rectangle, and then draw a value between $[0, a]$ at random as the width of the rectangle.

\textbf{Circle erasing}. Given an image of size $a\times a$, we first select an arbitrary point within the image as the center of the circle, and then draw a value between $[a/5,2a/5]$ at random as its radius.

\textbf{Edge erasing}. Given an image of size $a\times a$, we first select an arbitrary edge from left, right, top, bottom, and then draw a value between $[0,a/2]$ at random as the width of edge to erase.
    
    \subsection{Quantitative results}
    \label{sec:qr}
    In this section, we first report results on Human3.6M and compare with state-of-the-art methods in regular settings.
    Then we evaluate our method on incomplete images to test the robustness of 3D pose estimation.
    Some recent works estimate 3D pose by recovering 3D human mesh based on SMPL model~\cite{loper2015smpl}.
    We compare with these model-based methods on Human3.6M and 3DPW.
    Finally, we quantitatively evaluate our method on MPI-INF-3DHP without training on it, to evaluate the generalization  of our PONet.
 
    \subsubsection{Results on Human3.6M in regular settings}
        We first compare the proposed method with state-of-the-art 3D pose estimation methods in regular settings, \ie, all testing images are precisely cropped using ground truth bounding boxes so that the subjects always appear right in the center of testing images with all the joints within the images.

            \begin{table}[h]
                \centering
                \resizebox{0.8\linewidth}{!}{
                \begin{tabular}{lcc}
                    \toprule 
                    \textbf{Method (Authors)} & \textbf{MPJPE} & \textbf{PA-MPJPE} \\
                    \midrule
                    Martinez~\etal~\cite{martinez2017simple} &62.9  & 45.5 \\
                    Pavlakos~\etal~\cite{Pavlakos_2018_CVPR} &56.2  & 41.8 \\
                    Yang~\etal~\cite{Yang_2018_CVPR}        &58.6   &37.7   \\
                    Wang~\etal~\cite{Wang_2019_ICCV}        &52.6  &40.7 \\
                    Cai~\etal~\cite{Cai_2019_ICCV}  &50.6 & 40.2\\
                    Li~\etal~\cite{Li_2020_CVPR}    &50.9   &38.0   \\
                    Xu~\etal~\cite{Xu_2020_CVPR}   &49.2   &38.9   \\
                    \midrule
                    Ours &{56.1} &{39.8}\\
                    \bottomrule
                \end{tabular}
                }
                \caption{Quantitative results on Human3.6M in regular settings.}
                \label{tab:h36m_regular}
            \end{table}
            
        The results on Human3.6M in terms of MPJPE and PA-MPJPE are shown in Tab.~\ref{tab:h36m_regular}.
        All methods in this table, except for ours, regress the 3D joint locations so that they can fit the data by distorting the body.
        By contrast, our method uses a skeleton-shaped human model with fixed-length limbs so that our pose estimation will always satisfy anthropometric constraints, at the cost of some fitting accuracy.
        Even so, our results are on par with these methods.
            
        \subsubsection{Results on images with noisy bounding boxes } \label{sec:bbox}
\begin{figure}[htb]
            \centering
            \includegraphics[width=1\linewidth]{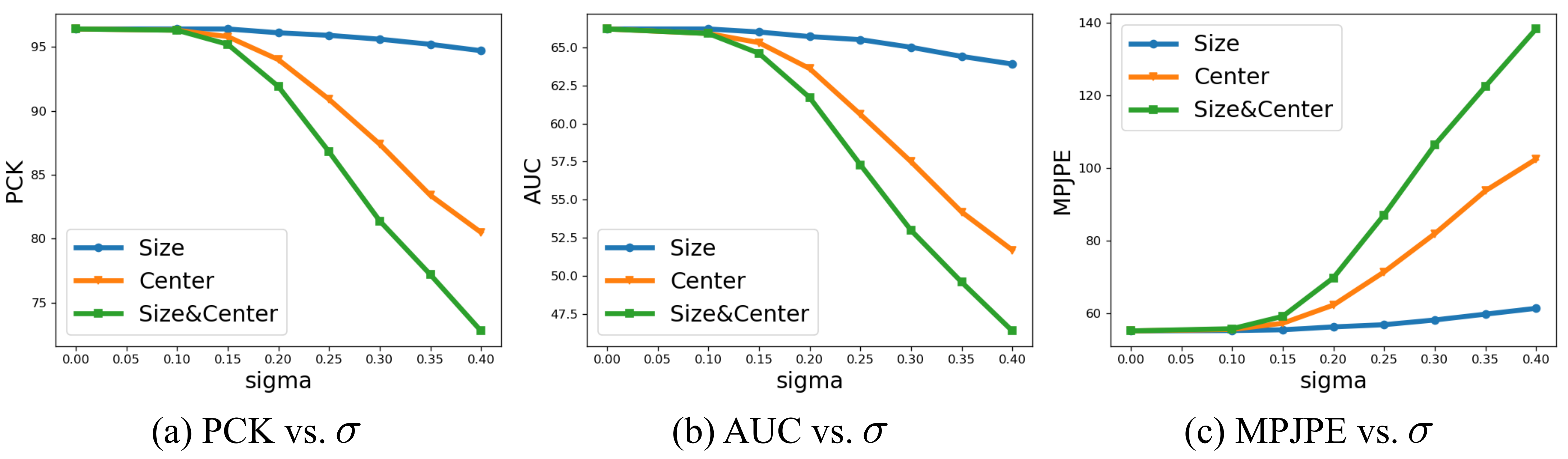}
            \caption{Results on Human3.6M with Gaussian noises $\mathcal{N}(0,\sigma^2)$ added to the centers or/and sizes of GT bounding boxes.}
            \label{fig:bbox}
        \end{figure}

We carry out experiments on Human3.6M by adding Gaussian noises to the centers~$C$ and/or the sizes~$S$ of ground truth bounding boxes,
        \begin{displaymath}
        \begin{array}{lc}
             C \leftarrow C + S\times\mathcal{N}((0,0), (\sigma^2_c,\sigma^2_c)),  \\
             S \leftarrow S + S\times\mathcal{N}(0, \sigma^2_s),
        \end{array}
        \end{displaymath}
where $\sigma_c$ and $\sigma_s$ control the level of noise relative to the GT size of bounding box.
The results are shown in Tab.~\ref{tab:bbox_noise} and Fig.~\ref{fig:bbox}.
        
        \begin{table}[h]
            \centering
            \resizebox{0.5\linewidth}{!}{
            \begin{tabular}{lrrr}
            \toprule
            {$(\sigma_c, \sigma_s)$}  &PCK   &AUC &MPJPE \\
            \midrule
            $(0,0)  $   &96.4   &66.2   &56.1 \\
            $(0.1,0)$   &96.3   &65.9   &56.6    \\
            $(0.15,0)$  &95.8   &65.3   &58.3    \\
            $(0.2,0)$   &94.0   &63.6   &63.3    \\
            $(0.25,0)$  &90.9   &60.6   &71.4    \\
            $(0.3,0)$   &87.4   &57.5   &81.9   \\
            $(0.35,0)$  &83.4   &54.2   &93.7   \\
            $(0.4,0)$   &80.5   &51.7   &102.4   \\
            \midrule
            $(0,0.1)$   &96.4   &66.2    &56.3\\
            $(0,0.15)$  &96.4   &66.0    &56.5\\
            $(0,0.2)$   &96.1   &65.7    &57.3\\
            $(0,0.25)$  &95.9   &65.5    &58.9\\
            $(0,0.3)$   &95.6   &65.0    &59.2\\
            $(0,0.35)$  &95.2   &64.4    &60.8\\
            $(0,0.4)$   &94.7   &63.9    &62.4\\
            \midrule
            $(0.1,0.1)$ &96.3   &65.9   &56.8\\
            $(0.15,0.15)$ &95.2 &64.6   &59.2\\
            $(0.2,0.2)$ &91.9   &61.7   &69.8\\
            $(0.3,0.3)$ &81.4   &53.0   &106.3\\
            $(0.4,0.4)$ &72.8   &46.4   &138.2\\
            \bottomrule
            \end{tabular}
            }
            \caption{Performance of our method on Human3.6M with different levels of Gaussian noises in the bounding boxes.}
            \label{tab:bbox_noise}
        \end{table}
        
        We can see that: (a) When $\sigma_c<0.1$ and $\sigma_s<0.1$, 
        the PCK and AUC of our approach drop less than $1\%$, while MPJPE drops about $1.2\%$, proving that our performance is very stable on images with small noise~(25.6 pixels) added to the bounding boxes.
        (b) When $\sigma_c=0.2$ and $\sigma_s=0.2$, the PCK only decreases $4.7\%$ and AUC decreases $6.8\%$, which shows that our PONet can work rather well even when both the size and the center of bounding boxes have an average error of $51.2$ pixels.
        The experiments demonstrate that our method does not rely on accurate bound boxes.
        This implies that in practical applications, we can use a person detector to obtain a rough bounding box to crop the image, without losing much accuracy in the performance of 3D human pose estimation.

        \subsubsection{Results on Human3.6M on incomplete images}
        Human3.6M is captured in monitored laboratory environment with precise bounding boxes and little noise.
        In practical applications, occlusion and truncation frequently occur in testing images.
        To find out the performance of 3D pose estimation in these challenging cases, it's not enough to only evaluate in ideal settings.
        To this end, we introduce three kinds of synthetic disturbance, including random object occlusion, random translation and random erasing to the testing images and evaluate the robustness of the proposed method on these images and compare it with several state-of-the-art keypoint-detection-based methods.
        To reduce the impact of random number generator, we independently repeat each experiment for 5 times, and report the average results in Tab.~\ref{tab:robustness}.
            \begin{table}[htb]
                \centering
                \resizebox{\linewidth}{!}{
                \setlength{\tabcolsep}{2pt}
                \begin{tabular}{lccccccc}
                    \toprule 
                    \multirow{2}{*}{\textbf{Method~(Authors)}} & \multirow{2}{*}{\textbf{None}}                &\multirow{2}{*}{\textbf{Occl.}} & \multicolumn{2}{c}{\textbf{Translation}} &\multicolumn{3}{c}{\textbf{Erasing}}\\
                    \cline{4-5}\cline{6-8}
                    {} & {}&{}& {25\%}&{40\%}&Rect.& Circle & Edge \\
                    \midrule
                    Martinez~\etal~\cite{martinez2017simple}   &62.9 &88.8 &121.2&261.7 &112.7 &87.6 &115.5 \\
                    Zhou~\etal~\cite{zhou2017towards}      &64.9 &90.0 &183.9&303.2 &109.6 &81.6 &117.5 \\
                    Wang~\etal~\cite{Wang_2019_ICCV}       &52.6 &88.6 &115.3&277.9 &108.1 &81.1 &109.3 \\
                    Li~\etal~\cite{Li_2020_CVPR}         &\bf50.9 &96.7&107.6&248.6&148.3&108.5&114.9\\
                    \midrule
                    {Ours before PC}    &56.2  &66.5    &60.5   &82.0 &67.5 &60.2 &71.4\\
                    {Ours after PC} &56.1  &\bf65.6 &\bf60.3 &\bf68.3&\bf66.2&\bf59.8 &\bf68.9\\
                    \bottomrule
                \end{tabular}
                }
                \caption{Comparison with state-of-the-art keypoint-detection-based methods on images from Human3.6M with missing joints or limbs in terms of MPJPE. PC denotes  pose complementation. 
                }
                \label{tab:robustness}
            \end{table}

        All the methods across Tab.~\ref{tab:robustness} show to have certain robustness to synthetic object occlusion~\cite{sarandi2018robust}, but our method proves to be more robust in this testing case.
        In terms of translation, our method turn out to have much better robustness than the other ones.
        While other methods already more than double their errors under $25\%$ translation, our MPJPE only increases $7.5\%$.
        Under $40\%$ translation, the other methods' MPJPEs sharply increase to more than $240$mm.
        By contrast, our result after pose complementation remains smaller than $69$mm, 
        demonstrating the outstanding robustness of our PONet on truncated images.
        We also randomly erasing a rectangle, circle or a certain edge on the input image to simulate keypoint and limb absent cases. Again our method outperforms the other ones, although we did not use data augmentation of erasing in training. 
        In summary, although our method does not achieve the best score in ideal settings, it performs much better on incomplete images including  occlusion, translation and erasing.

        There are four reasons why our method performs significantly better on images with missing joints or limbs.
        Firstly, our method does not detect the only intermittently visible body keypoints, but learn to estimate the limb orientation using a region-based representation.
        This pure orientation-based design allows us to estimate the 3D pose even when some limbs are partially invisible.
        Secondly, through pose complementation, our method can output a full-body pose estimation even if some limbs are completely out-of-image.
        Thirdly, we embed a fixed-length skeleton-shaped human model into the network so that the pose estimation will always satisfy anthropometric constrains such as limb ratios.
        Lastly, the 3D limb orientations are voted by all the pixels in the estimated orientation maps weighted with the corresponding confidence. This voting scheme can suppress the noise and outliers in the estimated 3D orientation maps, making the 3D orientation estimation and thus 3D pose estimation more robust and stable.
    
        \subsubsection{Comparison with SMPL-based methods}
        We compare with these SMPL-based methods because they also introduce anthropometric constraints to the estimated 3D poses.
        First, we compare our method with them on Human3.6M in regular settings. 
        The second column of Tab.~\ref{tab:hmr} shows the PA-MPJPE of our method and some recent ones
        Our method outperforms all the other approaches.
         \begin{table}[h]
                \centering
                \resizebox{1\linewidth}{!}{
                \begin{tabular}{lccc}
                    \toprule 
                    \multirow{2}{*}{\textbf{Method~(Authors)}} & \multirow{2}{*}{\textbf{H3.6M}} & {\textbf{H3.6M}} & \multirow{2}{*}{\textbf{3DPW}}\\
                    {}& {} &\textbf{(Occlusion)}&\\
                    \midrule
                    Bogo~\etal~\cite{bogo2016keep} &82.3 & 159.4& 114.0\\
                    Pavlakos~\etal~\cite{Pavlakos_2019_CVPR} &-&145.6 &151.3\\
                    Kanazawa~\etal~\cite{Kanazawa_2018_CVPR} &56.8 &82.2   &103.8\\
                    Kolotouros~\etal~\cite{Kolotouros_2019_CVPR} &50.1 &74.4 &104.8\\
                    Kolotouros~\etal~\cite{Kolotouros_2019_ICCV} &\underline{41.1} &64.9 &95.4\\
                    Zhang~\etal~\cite{Zhang_2020_CVPR} &41.7 &\underline{56.4} &\bf72.2\\
                    \midrule
                    {Ours} &\bf{39.8} &\bf{47.6} &\underline{76.2}\\
                    \bottomrule
                \end{tabular}
                }
                \caption{Comparison with state-of-the-art SMPL-based methods on Human3.6M and 3DPW in terms of PA-MPJPE. The best results are marked in bold and second best ones are underlined. No data from 3DPW are used to train our model.}
                \label{tab:hmr}
            \end{table}
        
        We also compare our method on datasets with occlusion. 
        The third column of Tab.~\ref{tab:hmr} reports results on Human3.6M with synthetic occlusions.
        Our method, again, outperforms all the other methods including a very recent work~\cite{Zhang_2020_CVPR} that specially designed for pose estimation on object-occluded images.
        We show the performance on 3DPW in the last column of Tab.~\ref{tab:hmr}.
        Our method achieves the second best performance on this dataset without training on it.
        
        In summary, our PONet performs better in both the original and occluded datasets, demonstrating the superiority of the proposed method over the others in terms of the accuracy of 3D joint location prediction and robustness.
    
            \begin{table}[h]
                \centering
                 \resizebox{1\linewidth}{!}{
                \begin{tabular}{lcccc}
                \toprule
                \textbf{Method~(Authors)} & \textbf{CE?} & \textbf{PCK} & \textbf{AUC} & \textbf{MPJPE} \\
                \midrule
                    Mehta~\etal~\cite{mehta2017monocular}   &N  &75.7   &39.3 &117.6    \\
                    Mehta~\etal~\cite{mehta2017vnect}       &N  &76.6   &40.4 &124.7    \\
                    Habibie~\etal~\cite{Habibie_2019_CVPR}  &N  &81.5   &44.5 &90.7     \\
                    Luo~\etal~\cite{luo2018orinet}          &N  &81.8   &45.2 &89.4     \\
                    \midrule
                    Luo~\etal~\cite{luo2018orinet}          &Y  &64.6   &32.1   &--     \\
                    Yang~\etal~\cite{Yang_2018_CVPR}        &Y  &69.0   &32.0   &--     \\
                    Zhou~\etal~\cite{zhou2017towards}       &Y  &69.2   &32.5   &137.1  \\
                    Habibie~\etal~\cite{Habibie_2019_CVPR}  &Y  &69.6   &35.5   &127.0 \\
                    Pavlakos~\etal~\cite{Pavlakos_2018_CVPR}&Y  &71.9   &35.3   &-- \\
                    Wang~\etal~\cite{Wang_2019_ICCV}        &Y  &71.9   &35.8   &--  \\
                    Kanazawa~\etal~\cite{Kanazawa_2018_CVPR}&Y  &72.9   &36.5   &124.2 \\
                     Li~\etal~\cite{Li_2020_CVPR}            &Y  &81.2   &46.1   &\bf99.7   \\
                \midrule
                    {Ours}~ &Y&\bf76.1 &\bf40.6 & 115.0 \\
                \bottomrule
                \end{tabular}
                }
                \caption{Results on MPI-INF-3DHP dataset.
                For PCK and AUC, higher is better; For MPJPE, lower is better. MPJPE is evaluated without Procrustes alignment.
                CE denotes cross-dataset evaluation without training on this dataset.}
                \label{tab:mpi_inf_3dhp}
            \end{table}
            
        \subsubsection{Evaluation of cross-dataset generalization}
        \begin{figure*}[tbh]
            \centering
            \includegraphics[width=\linewidth]{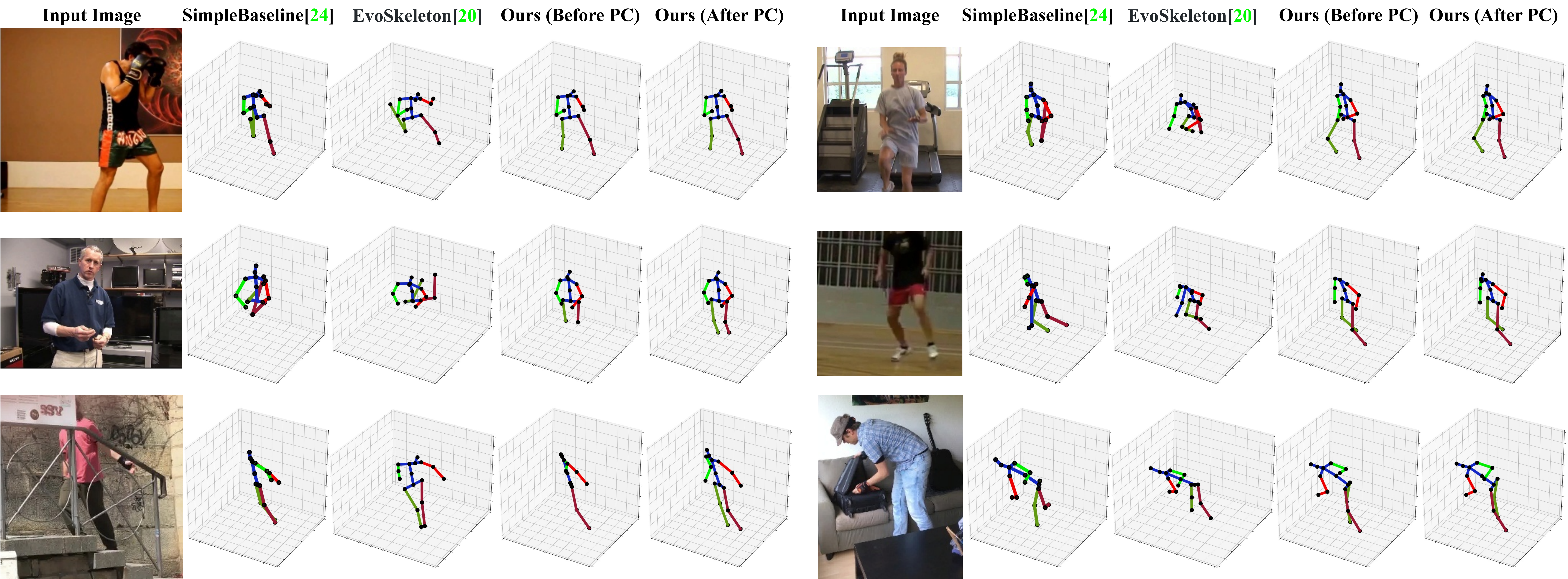}
            \caption{Qualitative results on  MPII~(first two rows) and 3DPW~(last row) and comparison  with \cite{martinez2017simple} and \cite{Li_2020_CVPR}. PC denotes \emph{pose complementation}. Our method is robust to joint absence~(first row), limb absence~(second row) and occlusion~(last row).}
            \label{fig:qual_comparison}
        \end{figure*}
        
            To quantitatively show the generalization on unseen images, we compare with other methods on MPI-INF-3DHP without using any data from this dataset for training. 
            In addition to MPJPE, we report Percentage of Correct Keypoints~(PCK) thresholded at 150mm and Area Under the Curve~(AUC) over a range of PCK thresholds. 
       
            The results are shown in Tab.~\ref{tab:mpi_inf_3dhp}. Our method outperforms state-of-the-art methods by a large margin without training on this dataset, demonstrating our method's excellent generalization to unseen images.

    \subsection{Qualitative results}
        In this section, we qualitatively compare our method with state-of-the-art methods on MPII and 3DPW.
        In Fig.~\ref{fig:qual_comparison}, we show six examples with invisible parts  and compare with two approaches that based on 2D keypoint detection, SimpleBaseline~\cite{martinez2017simple} and a very recent work EvoSkeleton~\cite{Li_2020_CVPR}.
        SimpleBaseline uses \emph{StackedHourglass}~\cite{newell2016stacked} to detect 2D poses, while EveSkeleton uses a better 2D pose estimator \emph{HRNet}~\cite{Sun_2019_CVPR}, and achieved state-of-the-art results on Human3.6M in regular settings.

        The first row of Fig.~\ref{fig:qual_comparison} shows two cases from MPII with only a single out-of-image joint. Neither of SimpleBaseline and EvoSkeleton can correctly predict the 3D location of the missing joint, while our method provides visually appealing results even without pose complementation.
        The second row of Fig.~\ref{fig:qual_comparison} shows two cases with invisible limbs.
        The performance of the two keypoint-detection-based methods drops drastically,  compared to single joint absent cases.
        By contrast, our method can correctly estimate the configuration of the visible parts, and provide a reasonable full-body pose estimation after pose complementation.
        The third row of Fig.~\ref{fig:qual_comparison} shows two occluded images.
        Again, our method performs better in these cases as well.
        
        
        

        \subsection{Failure cases}
\begin{figure*}[htb]
            \centering
            \includegraphics[width=\linewidth]{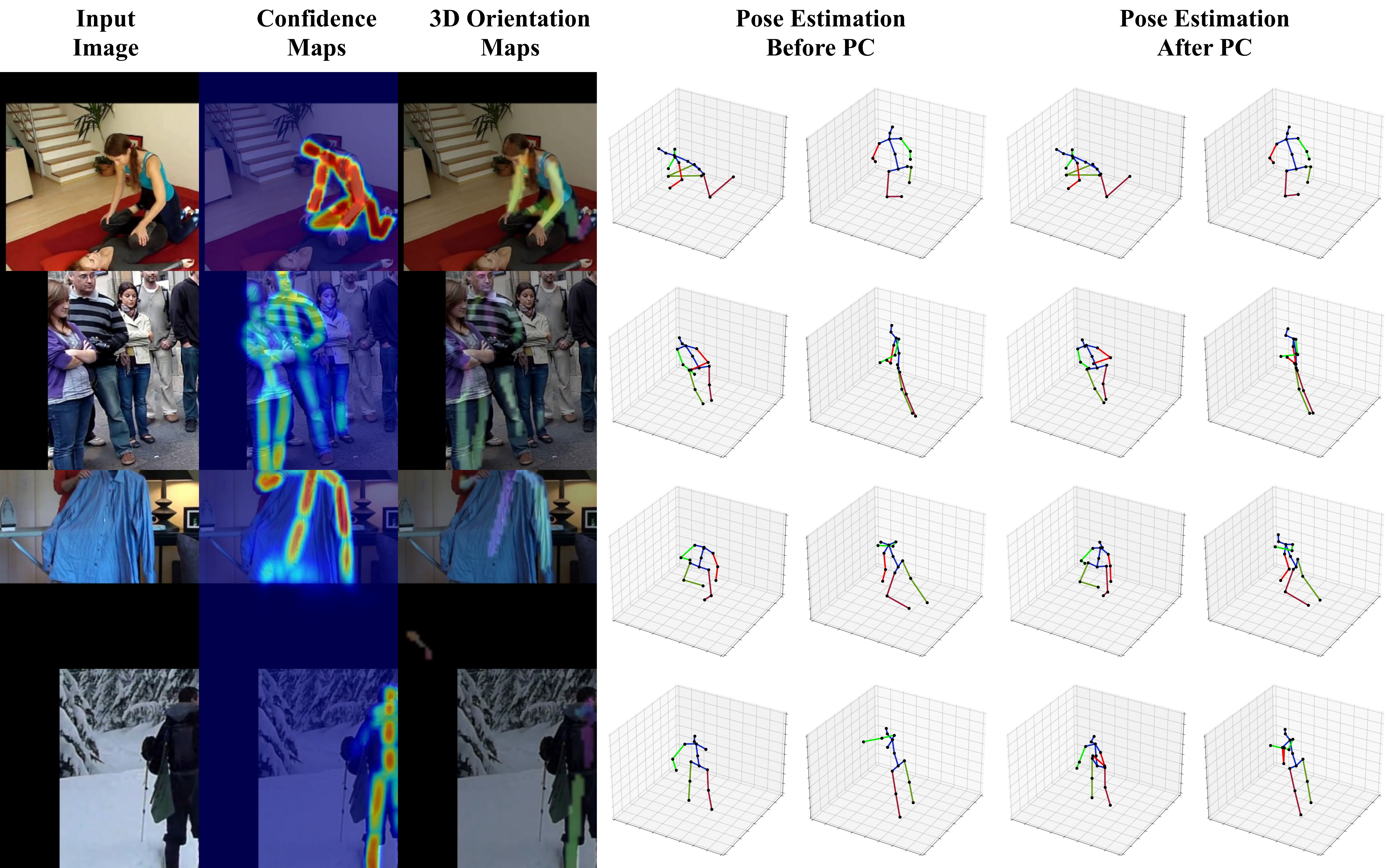}
            \caption{Failure Cases.}
            \label{fig:failure_cases}
        \end{figure*}

    In Fig.~\ref{fig:failure_cases}, we provide 4 typical failure cases.
    The first two rows of Fig.~\ref{fig:failure_cases} show 
    two failure cases in limb  region estimation in multi-person scenarios.
    The third row of Fig.~\ref{fig:failure_cases} is a false positive case on clothes.
    The last row of Fig.~\ref{fig:failure_cases} shows an example 
    that the right half of the body is truncated so 
    that the network gets confused about the left and right side of the body.

    \subsection{Ablation study}
    To analyze the effectiveness of each component in our method, we conduct ablation study on Human3.6M in both regular setting and translation up to $40\%$ of the image size.
    In Tab.~\ref{tab:ablation}, \emph{Baseline} refers to model trained with $\mathcal{L}_{CM}$,  $\mathcal{L}_{OM2D}$ and $\mathcal{L}_{OM3D}$.
    \emph{Aug.} refers to using the data augmentation of synthetic occlusion and $40\%$ random translation.
    \emph{PC} refers to pose complementation.
    
    \begin{table}[h]
        \centering
        \resizebox{\linewidth}{!}{
        \begin{tabular}{lcc}
            \toprule 
            \textbf{Method} &\textbf{Regular} &\textbf{Translation} \\
            \midrule
             Baseline & 59.8  & 92.2\\
             Baseline + $\mathcal{L}_{P3D}$ &58.4 & 87.7\\
             Baseline + $\mathcal{L}_{P3D}$  + Aug. &56.2 & 82.0\\
             Baseline + $\mathcal{L}_{P3D}$  + Aug. + PC & 56.1 & 68.3\\
            \bottomrule
        \end{tabular}
        }
        \caption{Ablation study on Human3.6M in terms of MPJPE.}
        \label{tab:ablation}
    \end{table}
    
    From Tab.~\ref{tab:ablation} we can see that, the baseline yields MPJPE of 92.2 on images with $40\%$ translation, outperforms all the other methods in Tab.~\ref{tab:robustness}, proving the  superiority of our orientation-based method over keypoint-detection-based ones in improving the robustness.
    End-to-end training can improve the performance on both complete and incomplete images, demonstrating the importance of our differentiable orientation extraction.
    Data augmentation of occlusion and large translation can further improve the performance.
    At last,  the performance gain brought by pose complementation in regular setting is limited, but it can improve the performance on translation of $40\%$ by a large margin.

\section{Conclusion}
 In this paper, we propose PONet, a robust and effectual 3D human pose estimation 
 network based on learning orientation only.
 Our method bypasses 2D keypoint detection, which is prone to errors in the absence of body parts, by learning three sets of maps that encode the limb confidence, 2D and 3D orientations of each limb.
 This design allows us to predict the 3D limb orientations on images with absent joints.
 In the more challenging scenarios where limbs are completely occluded or out-of-image,
 PONet can provide a complete 3D pose estimation by inferring the 3D orientations of the invisible limbs from the visible ones using pose complementation.
 We evaluate our method on multiple datasets including Human3.6M, MPII, MPI-INF-3DHP and 3DPW.
 Extensive experiments demonstrate the gratifying  robustness of the proposed method.

{\small
\bibliographystyle{ieee_fullname}
\bibliography{egbib}
}

\newpage



\bf{If you will not include a photo:}
\begin{IEEEbiographynophoto}{John Doe}
Use $\backslash${\tt{begin\{IEEEbiographynophoto\}}} and the author name as the argument followed by the biography text.
\end{IEEEbiographynophoto}

\vfill

\end{document}